\documentclass[10pt,twocolumn,letterpaper]{article}

\usepackage{iccv}      
\usepackage{pifont}
\newcommand{\greencheck}{\textcolor{green}{\textcircled{\ding{51}}}}
\newcommand{\redcross}{\textcolor{red}{\textcircled{\ding{55}}}}

\usepackage{amsmath}
\usepackage{amsthm}
\usepackage{booktabs}
\usepackage{algorithm}
\usepackage{algorithmic}
\usepackage{algorithm}
\usepackage{algorithmic}
\usepackage{enumerate} 
\usepackage{multirow}
\usepackage{color}
\usepackage{tikz}
\usepackage{booktabs}
\usepackage{pifont}
\usepackage{marvosym}
\usepackage{booktabs}
\usepackage{bm}
\usepackage{newtxtext}
\usepackage{newtxmath}
\usepackage{subcaption}
\usepackage{multirow}
\usepackage{amsmath}
\usepackage{array}
\usepackage{threeparttable}
\usepackage{makecell}
\usepackage{mathrsfs}
\usepackage{xcolor}
\usepackage{pifont}
\usepackage{newfloat}
\usepackage{listings}

\definecolor{iccvblue}{rgb}{0.21,0.49,0.74}
\usepackage[pagebackref,breaklinks,colorlinks,allcolors=iccvblue]{hyperref}

\def\paperID{3708} 
\def\confName{ICCV}
\def\confYear{2025}

\title{LLM-enhanced Action-aware Multi-modal Prompt Tuning for Image-Text Matching}

\author{
Mengxiao Tian\textsuperscript{1,2,3}, Xinxiao Wu\textsuperscript{1,2}, Shuo Yang\textsuperscript{2} \thanks{Corresponding author}\\
\textsuperscript{1} Beijing Key Laboratory of Intelligent Information Technology, \\ School of Computer Science \& Technology,  Beijing Institute of Technology, China \\ 
\textsuperscript{2} Guangdong Laboratory of Machine Perception and Intelligent Computing, \\ Shenzhen MSU-BIT University, China \\ 
\textsuperscript{3} Beijing Research Center of Intelligent Equipment for Agriculture, China\\
}

\begin{document}
\maketitle
\begin{abstract}
		Driven by large-scale contrastive vision-language pre-trained models such as CLIP, recent advancements in the image-text matching task have achieved remarkable success in representation learning. Due to image-level visual-language alignment, CLIP falls short in understanding fine-grained details such as object attributes and spatial relationships between objects. Recent efforts have attempted to compel CLIP to acquire structured visual representations by introducing prompt learning to achieve object-level alignment. While achieving promising results, they still lack the capability to perceive actions, which are crucial for describing the states or relationships between objects. Therefore, we propose to endow CLIP with fine-grained action-level understanding by introducing an LLM-enhanced action-aware multi-modal prompt-tuning method, incorporating the action-related external knowledge generated by large language models (LLMs). Specifically, we design an action triplet prompt and an action state prompt to exploit compositional semantic knowledge and state-related causal knowledge implicitly stored in LLMs. Subsequently, we propose an adaptive interaction module to aggregate attentive visual features conditioned on action-aware prompted knowledge for establishing discriminative and action-aware visual representations, which further improves the performance. Comprehensive experimental results on two benchmark datasets demonstrate the effectiveness of our method. 
\end{abstract}    
\section{Introduction}
\label{sec:intro}

Image-text matching has gained much attention in recent years, aiming to match images with the most relevant texts and vice versa. This is a fundamental task in vision and language research, involving multimodal reasoning and alignment of visual and textual concepts at different levels, such as cross-modal retrieval~\cite{pan2023fine,ge2023cross}, natural language visual reasoning~\cite{gupta2023visual,chen2024large}, and visual question answering~\cite{ravi2023vlc,guo2023images}. 
	\begin{figure}
		\centering
		\begin{subfigure}[b]{0.22\textwidth}  
			\centering
			\includegraphics[width=\textwidth]{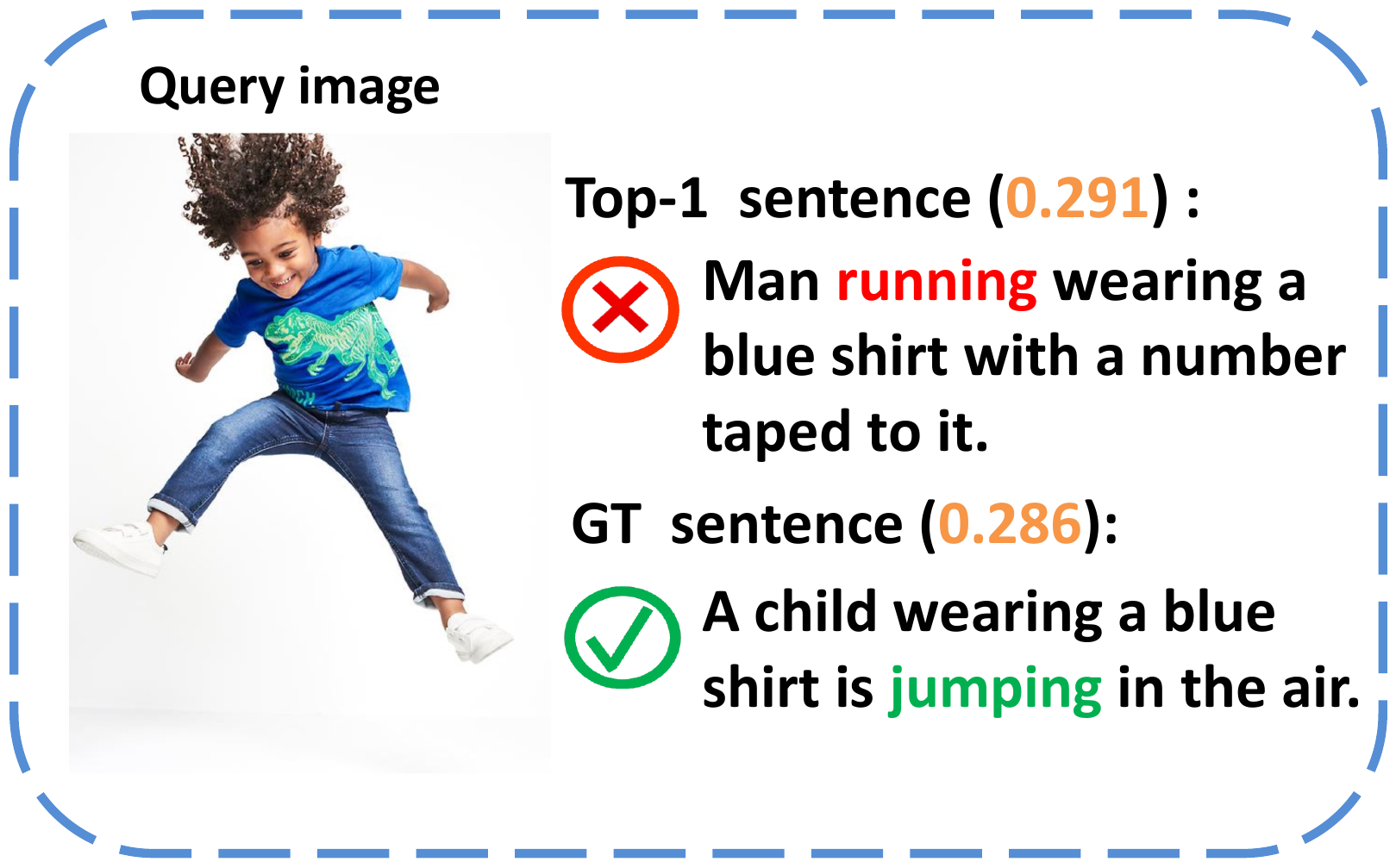}
			\caption{Image-to-Text Matching}
			\label{subfig:image_to_text}
		\end{subfigure}
		\hspace{0.02\textwidth}  
		\begin{subfigure}[b]{0.22\textwidth}  
			\centering
			\includegraphics[width=\textwidth]{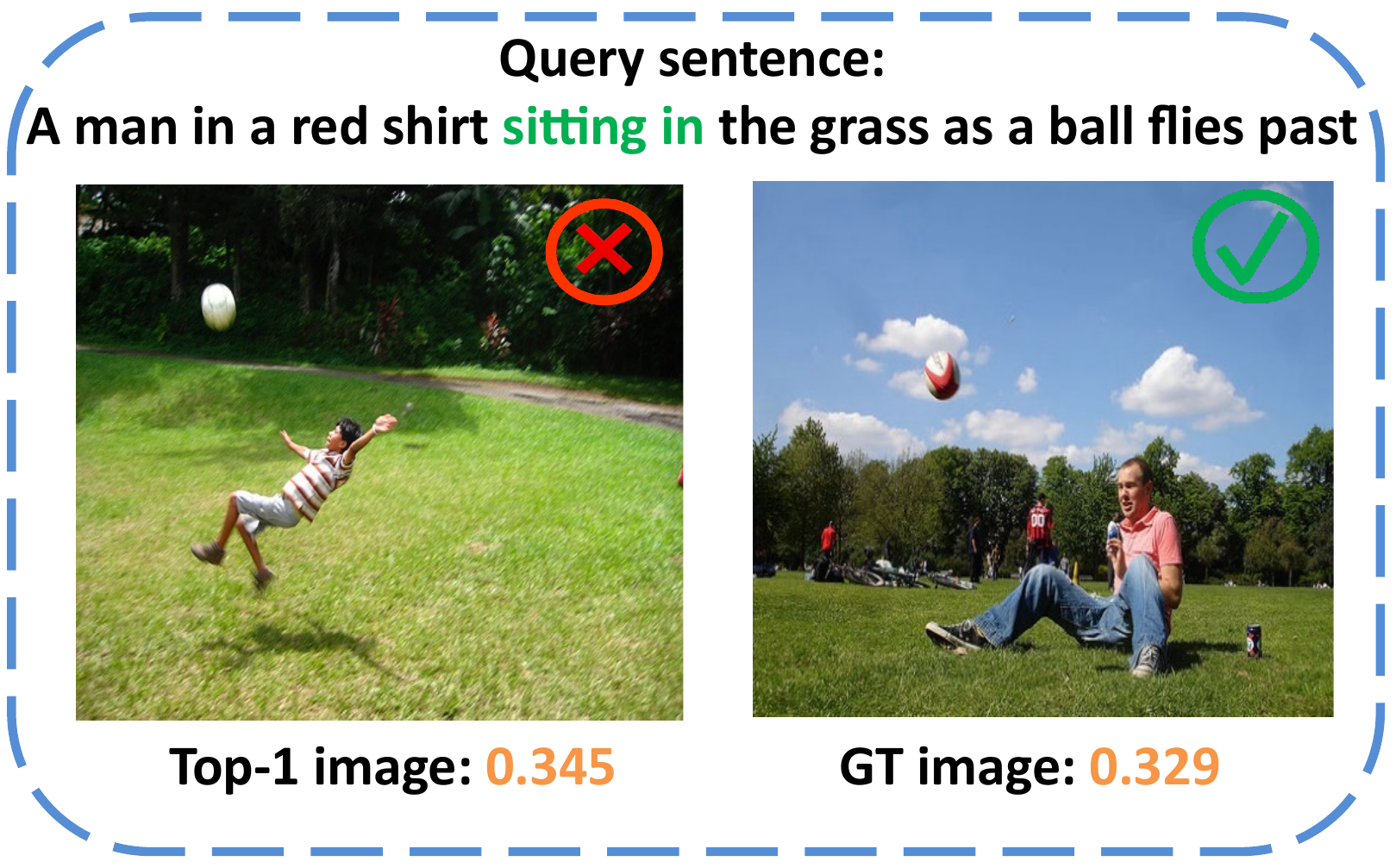}
			\caption{Text-to-Image Matching}
			\label{subfig:text_to_image}
		\end{subfigure}
        \caption{Failure cases of image-text matching (image-to-text and text-to-image) using CLIP on the COCO test set. Similarity scores are shown in orange, and correct matches are marked in {\protect\greencheck} and mismatches in {\protect\redcross}. (a) CLIP incorrectly retrieves the top-1 text for a given image query; (b) CLIP incorrectly retrieves the top-1 image for a given text query.}
        \vspace{-10pt}
		\label{diyiye}
	\end{figure}
	\textcolor{black}{Recently, large-scale pre-trained models such as CLIP~\cite{radford2021learning} have shown superior generalization and transferability on various downstream tasks~\cite{wang2021actionclip,zhu2022relclip,DBLP:conf/iclr/GengYT0Z23}, making CLIP one of the most widely used pre-trained models for image-text matching.}
	
	However, a large number of studies have demonstrated that discarding fine-grained visual information results in subpar performance on downstream tasks involving localization~\cite{zhong2022regionclip}, counting~\cite{paiss2023teaching}, and understanding relationships between objects or object attributes~\cite{zhu2022relclip,huang2024structure}.
	Recent studies~\cite{kan2023knowledge,wang2023position,liu2024multi} focus on object-level understanding by introducing learnable prompts that describe more details of image regions, \textit{i.e.}, objects or object attributes, 
	enabling CLIP to learn fine-grained representations, thereby enhancing fine-grained understanding. 
	Despite the impressive performance, whether CLIP can effectively perceive actions (\textit{i.e.}, states or relationships between objects) in the image-text matching task remains unresolved.
	
\begin{figure}
		\centering
		\includegraphics[width=0.8\linewidth]{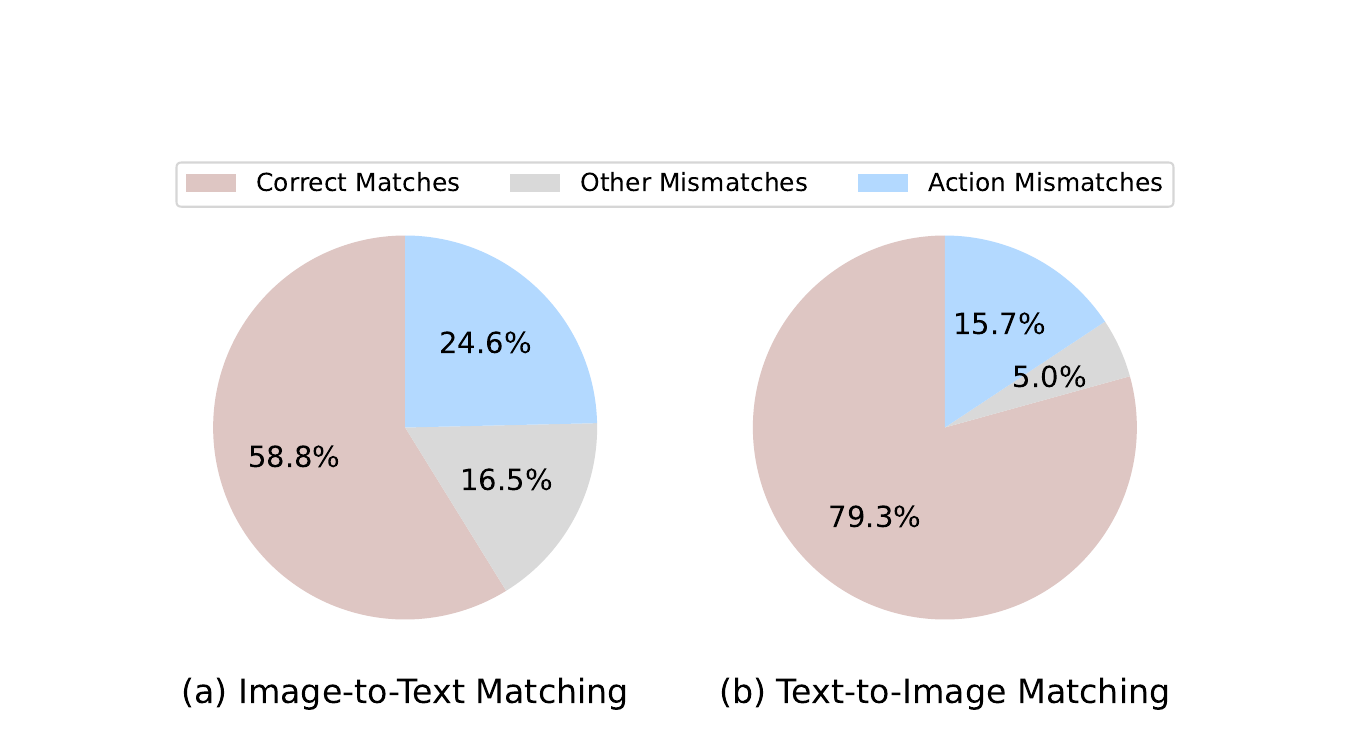}
		\caption{The statistical analysis of inconsistent actions between the query and the candidate in image-text matching using CLIP on the Flickr30K test set.}
        \vspace{-10pt}
		\label{tongji}
	\end{figure}

    \textcolor{black}{As shown in Figure~\ref{diyiye}, the CLIP score of the ground-truth image-text pair is lower than that of a mismatched pair. We further perform a statistical analysis of the action inconsistency between the query and the candidate, as shown in Figure~\ref{tongji}.} These results indicate that CLIP exhibits inadequacies in accurately perceiving actions.
	Therefore, we ask: \textit{How to endow CLIP with the fine-grained and action-aware perception to improve its image-text matching capability?}
	This paper aims to approach this from both textual and visual perspectives.
	
	From the textual perspective, we consider the compositionality and causality of actions. We assume that each action has  an initiator and a recipient, 
	and there is a logical relationship between the two regarding the specific action.
	Therefore, we propose to leverage the world knowledge about actions in LLMs to decompose the original textual descriptions into all possible action triplets in the format of  \textless subject, action, object\textgreater. Each triplet captures a pair of potential entities with their action interrelations. For example, as shown in Figure~\ref{diyiye}, the original description “A man wearing a blue
	shirt is jumping in the air” is decomposed into “\textless man, jumping, air\textgreater”. Furthermore, we believe that each action has its state, called action state, which can be caused or inferred by specific entities. Each action state provides more details and comprehensive descriptions to supplement the information contained in the action triplet. Thus, for each action triplet, we feed the hand-crafted instructions into LLMs to generate state descriptions about the action in the form of language prompts. For example, the action state “lifting both feet off the ground and propelling the body upwards” corresponds to the action triplet “\textless man, jumping, air\textgreater”. 
	
	From the visual perspective, we also need to consider how to embed the action knowledge obtained from the textual descriptions into model training, that is, how to use action triplets and action states to guide the image-text training of CLIP. To this end, we design action-aware prompts based on different action information during training: action triplet prompts and action state prompts, which will be fed into the image encoder of CLIP to enable its visual representation to better learn action information at different levels. The action triplet prompts guide the image encoder to capture the rich compositional semantics existing in the action relationships between entities, while the action state prompts guide the image encoder to pay more attention to the resulting states of action in the visual content.
	Furthermore, to reduce the disturbance of irrelevant or noisy action information in the action-aware prompts, we propose an adaptive interaction module, which adaptively attends to the salient action cues relevant to the visual content, facilitating fine-grained alignment with the corresponding textual representations.
	
	The main contributions are summarized as follows: (1) To the best of our knowledge, we are the first to enhance CLIP with fine-grained action-aware perception,  and propose an LLM-enhanced action-aware multimodal prompt tuning method that incorporates external action knowledge from LLMs to improve the image-text matching performance. (2) We design two complementary action-aware prompts that use fine-grained action knowledge from compositionality and causality perspectives, and introduce an action-aware adaptive interaction module to aggregate attentive visual features based on action knowledge. (3) Extensive experiments on public datasets show that our method achieves significant performance improvements over existing methods.
	
	\begin{figure*}
		\centering
		\includegraphics[width=0.93\textwidth]{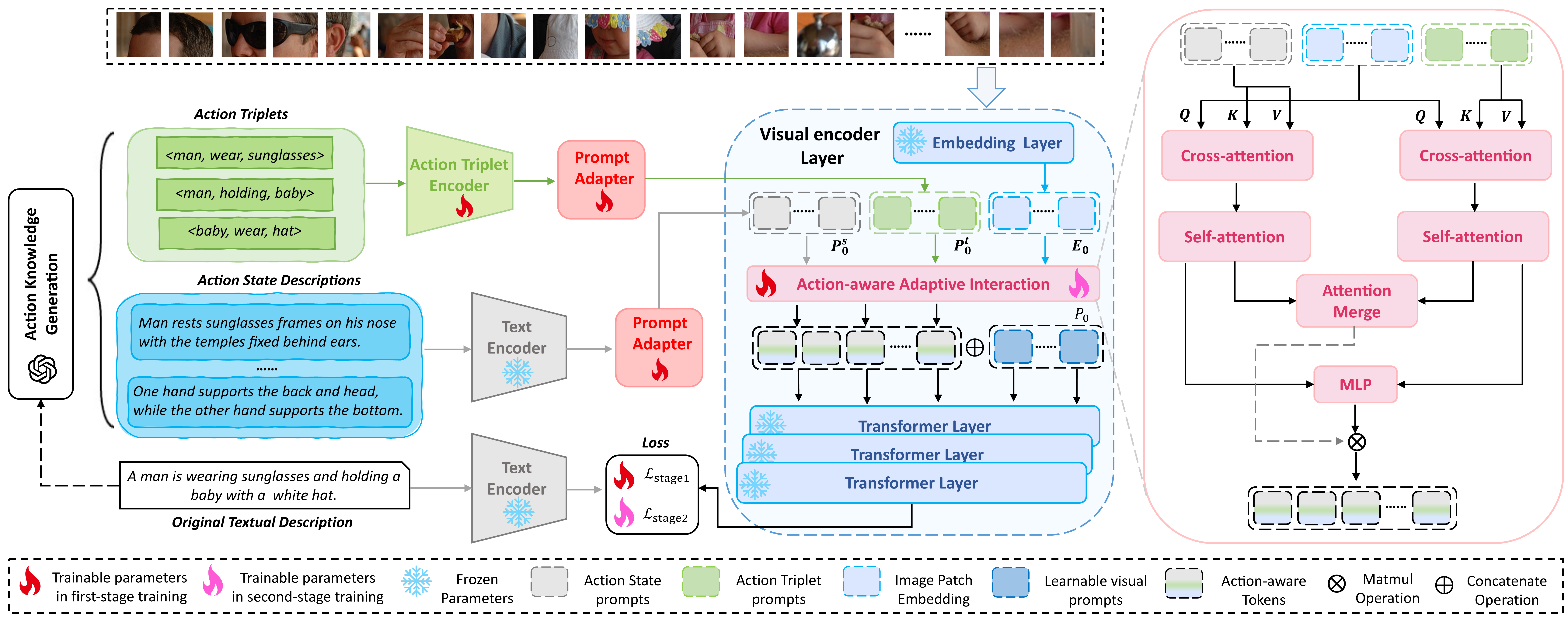}
		\caption{Overview of the proposed method. 
	}
	\label{pipeline}
    \vspace{-5pt}
	\end{figure*}
	\section{Related Work}
	\subsection{Image-Text Matching}
	Image-text matching has attracted much attention in recent years, and most existing methods attempt to embed images and texts into a common space to measure their semantic similarity. These methods can be categorized into coarse-grained and fine-grained matching.
	The coarse-grained matching methods~\cite{DBLP:conf/bmvc/FaghriFKF18,zheng2020dual,li2022image} utilize separate encoders to extract global features from images and texts, and then project them into a common embedding space. 
	The fine-grained matching methods~\cite{lee2018stacked,li2022action,ge2023cross,DBLP:journals/ijon/WuWCLW23} focus on learning fine-grained alignments between image regions and textual words. SCAN~\cite{lee2018stacked} proposes a stacked cross-attention network to explore fine-grained correspondences between image regions and words, inspiring a range of various sophisticated methods~\cite{qu2021dynamic,cheng2022cross,pan2023fine}. 
	Recently, many works~\cite{wang2020cross,diao2021similarity,fu2023learning} focus on employing GCN-based networks to learn local region-word alignment, achieving significant performance improvements. 
	Despite the significant progress achieved by exploring diverse fine-grained interaction patterns, these fine-grained methods still heavily rely on expert experience to handcraft networks.
	
	The most relevant method to our work is AME~\cite{li2022action}, which enhances action-aware representations using the action-similar texts from the memory bank. In contrast, we exploit the powerful in-context learning capability of LLMs to generate fine-grained action knowledge instead of relying solely on fixed original textual descriptions, thus enabling the model to achieve  more fine-grained action-aware visual understanding.
	
	\subsection{Prompt Learning for CLIP}
	
	Various methods for prompt learning~\cite{khattak2023maple,wang2023position,wang2024vilt}
	have emerged as an effective and parameter-efficient approach to enhance the adaptation of CLIP  to the downstream tasks, such as few/zero classification~\cite{zheng2023large,li2023cross}, object detection~\cite{long2023fine,li2024learning}, and video understanding~\cite{wang2021actionclip,wang2024vilt}.
	
	A method closely related to ours is actionCLIP~\cite{wang2021actionclip}, which leverages CLIP's joint visual-textual embedding space for action recognition. However, it only uses action names as text prompts with prefixes and suffixes that carry limited semantic meanings. In contrast, we adopt a systematic approach that uses LLM as a knowledge engine to generate knowledge-rich prompts tailored to actions. These prompts incorporate fine-grained knowledge from compositionality and causality perspectives to provide fine-grained guidance for representation learning. In addition, we introduce an additional module to maximize the  capture of action-related information.
	
	\subsection{External Knowledge from LLMs}
	Recently, the integration of task-relevant external knowledge from LLMs into prompts has been explored to enhance the discriminative ability of the prompt tuning process.
	KgCoOp~\cite{yao2023visual} constrains the learnable prompt embeddings with general knowledge to improve generalization. 
	Kan \textit{et al.}~\cite{kan2023knowledge} developed knowledge-aware prompts to assist the model in identifying the correspondences between images and different categories. Wang \textit{et al.}~\cite{wang2023learning} proposed to generate diverse forms of linguistic knowledge and conduct hierarchical prompt tuning for better performance. Menon \textit{et al.}~\cite{DBLP:conf/iclr/MenonV23} proposed to classify images by querying VLMs with descriptive features extracted from LLMs for interpretable and adaptable recognition. Su \textit{et al.}~\cite{su2023language} leveraged LLMs to extract causal commonsense knowledge for reasoning about only the “why” questions in video understanding. Furthermore, Tan \textit{et al.}~\cite{tan2023compound} proposed category-wise and content-wise guidance for prompt optimization, helping the model understand general category information while capturing intraclass variation. Buettner \textit{et al.}~\cite{buettner2024incorporating} distinguished visual descriptions of the same object class across different geographical regions by using LLMs. Niu \textit{et al.}~\cite{DBLP:conf/iclr/NiuG00C24} proposed a state-aware reasoning framework with LLM-generated knowledge for dynamic procedure planning in instructional videos.
	
	The above LLMs-based methods typically focus on general descriptions of entities and attributes in the text space. In contrast, we propose leveraging external action knowledge from LLMs to concentrate on the action states and relationships between objects. This knowledge can then be integrated into the prompts learning to enhance CLIP's fine-grained visual perception of actions.
	
	\section{Our Method}
	\subsection{Overview}
	In this paper, we propose an LLM-enhanced action-aware multi-modal prompt-tuning method, which incorporates external action knowledge from LLMs to enhance action-aware perception, with a two-stage training strategy.
	As illustrated in Figure~\ref{pipeline}, starting with the original text description, the action knowledge generation module extracts and generates two kinds of knowledge: action triplets that outline the relationships between entities, and action state descriptions that detail the states of entities related to the action. These forms of knowledge serve as prompts to enhance the action-aware perception abilities of CLIP. Furthermore, these prompts are integrated into the proposed action-aware adaptive interaction modules together with visual tokens to effectively capture various levels of visual information related to the action.
	
	\begin{figure*}
	\centering
	\includegraphics[width=0.9\textwidth]{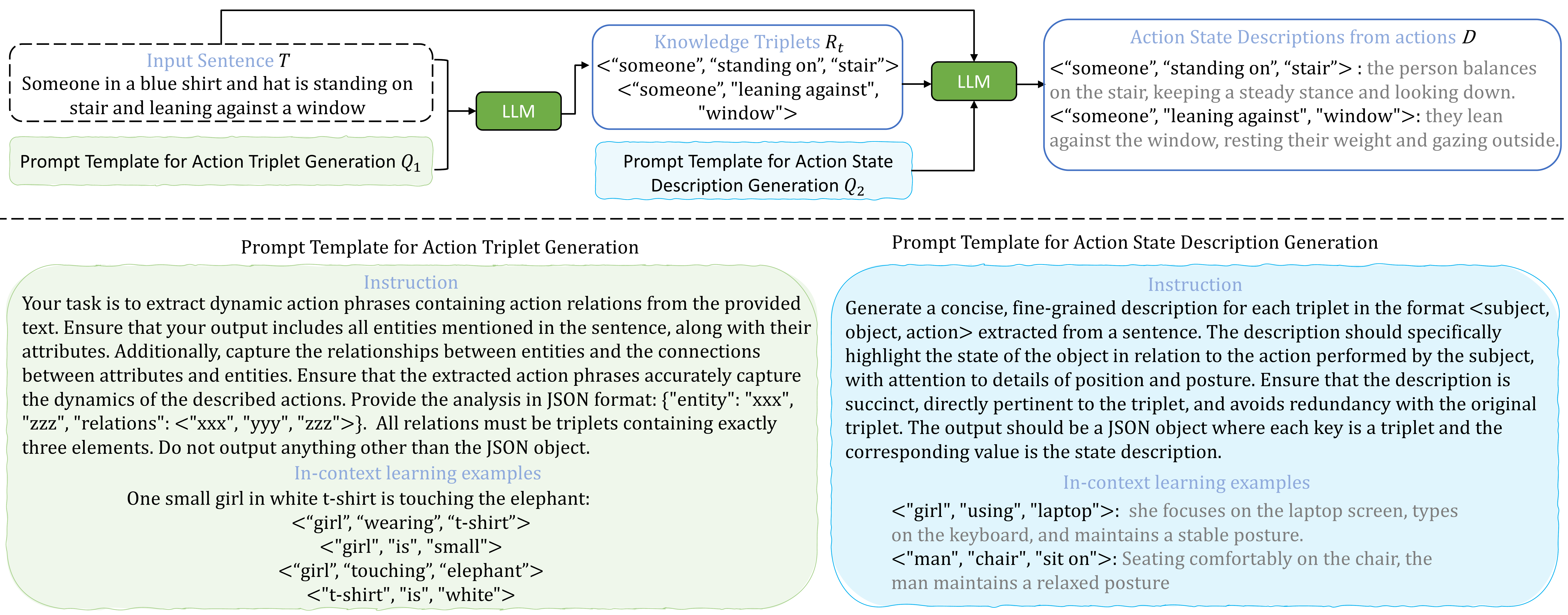}
	\caption{Action knowledge generation using LLM.}
	\label{fig:generate}
    \vspace{-10pt}
	\end{figure*}
	
	\subsection{Generation of Action Knowledge}
	\label{sec:generate}
	Figure~\ref{fig:generate} shows the generation of action knowledge by using the powerful in-context learning capability of LLMs. Our prompt template is generated using GPT-3.5, and minor adjustments do not affect the output. This knowledge includes not only fine-grained action relationships but also precise descriptions of their action states. Specifically, we design a prompt to instruct GPT-3.5 to construct action triplets, denoted as $\mathcal{R}_{t}$, and we further feed them with another instruction to generate knowledge-rich and fine-grained action state descriptions $D$ about actions in the form of language, 
	serving as valuable informative supplements for $\mathcal{R}_{t}$. 
	A simple format check is performed afterward to ensure the output meets the desired specifications. 
	It is important to note that the prompts can vary depending on datasets to accommodate different distributions of data. 
	
	\noindent \textbf{Action Triplet Generation.} We use one of the most powerful LLMs, GPT-3.5 to generate action triplet. Specifically, we first prepare several in-context learning examples of possible inputs along with their expected output triplets for GPT-3.5 to perform in-context learning. Each in-context sample consists of a textual description and corresponding action triplets. By combining multiple in-context samples, the LLM is constrained to generate higher-quality outputs while reducing irrelevant content.
	Subsequently, we use a well-designed instruction template $Q_{1}$ with in-context samples to instruct GPT-3.5 to parse an input caption  \( T \) into decoupled textual action triplets, denoted as $\mathcal{R}_{t}$, which include entities and their interrelated action relationships, formulated as \ 
	\begin{equation}     
	\mathcal{R}_{t} = \text{LLM} ([Q_{1}, T]).
	\end{equation} 
	Here \( \mathcal{R}_{t} = \{r_i\}_{i=1}^{N}\), and  \(r_i\) is represented as
	\begin{equation}
	r_i = \left\{ (e_i, r(e_i, e_j), e_j) \mid 1 \leq i, j \leq N \right\},
	\end{equation}
	where \( N \) denotes the valid number of entity pairs in \( T \), and \( r(e_i,e_j) \) is the action relationships between entity pair \( (e_i,e_j) \). Notably, our action triplets can contain entities along with their fine-grained attributes (\textit{e.g.,} {“(girl, is, small)”}), since actions inherently depend on both motion dynamics and static appearance. This ensures a comprehensive, contextually nuanced representation while maintaining focus on dynamics.
	
	\noindent \textbf{Action State Description Generation.}
	Relying solely on action triplets may not be sufficient to capture the factual details of actions due to their inherent ambiguity and variability across different scenarios. Therefore, 
	we consider the state of each action, referred to as action states, which can be caused or inferred by specific actions. Each action state contains more detailed and comprehensive descriptions, enabling the model to gain deeper insights into the fine-grained action semantics. 
	As shown in Figure~\ref{fig:generate}, we adopt a refined language instruction $Q_{2}$ with in-context samples to query GPT-3.5 to generate state descriptions about actions in the form of language, denoted as $D$, where the in-context samples consist of the action triplets, the original textual descriptions, and the expected output descriptions, formulated as 
	\begin{equation}     
	D = \text{LLM}(Q_{2}, \mathcal{R}_{t}).
	\end{equation}
	For example, when querying the GPT-3.5 about {\textit{“(girl, laptop, study)”}}, GPT-3.5 would generate an answer like {\textit{“She focuses on the laptop screen, types on the keyboard, and maintains a stable posture”}.}
	
	\subsection{Action-aware Multi-modal Prompt Tuning}
	\label{sec:prompt}
	
	To address this limitation of pre-trained models in action perception, we propose an action-aware multi-modal prompt tuning. Specifically, we design two complementary action-aware prompts based on different action information: action triplet prompts and action state prompts, which will be fed into the image encoder to help CLIP's visual representation better capture different levels of action information. 
	For the early layers of the image encoder, we map the generated prompted action knowledge into visual prompts by employing a lightweight action-aware prompt adapter, meaning that both types of action-aware prompts guided the encoding process of images. 
	Moreover, for the later layers of the image encoder, each layer utilizes an independent prompt, allowing the layer to independently capture distinct visual and semantic features of the image. 
	To further improve fine-grained visual perception, we propose an action-aware adaptive interaction module to adaptively capture the salient action cues relevant to the visual contents.
	
	\noindent \textbf{Action Triplet Prompts.} To make action triplet prompts carry sufficient compositional semantics present in actions, we design an action-aware triplet encoder with transformer layers to explicitly model the action relations between paired objects. Specifically, for each action triplet $(e_i, r(e_i, e_j), e_j)$ in $\mathcal{R}_{t}$, we generate the corresponding word embedding $W_y$ by employing a CLIP-based tokenizer together with a vocabulary embedding, formulated as
	\begin{equation}
	\textbf{W}_y = \text{WordEmb}_{\text{CLIP}}(y), \quad y \in \{e_i, r(e_j), e_j\}.
	\end{equation}
	Subsequently, we concatenate these embeddings to form a unified representation for each triplet, denoted as ${e}_{\Delta_i}$. This process involves handling a total of $K$ action triplets to produce $K$ semantic embeddings, which are subsequently fed into 
	multiple transformer layers to generate continuous prompt embedding features $\textbf{p}_{t}$,  formulated as
	\begin{equation}
	\textbf{{p}}_{t} = \text{Multi-Transformer}([{e}_{\Delta_1}, {e}_{\Delta_2},..,{e}_{\Delta_K}]).
	\end{equation}
	
	\noindent \textbf{Action State Prompts.} 
	Action state prompts can be constructed by plugging the generated action state descriptions from actions in a predefined prompt template like \textit{“{Fine-grained state description of action triplet \textless subject, action, object\textgreater is [description]}”}. These prompts guide the image encoder to focus on the resulting states of actions in visual content. We use a frozen text encoder to encode these state prompts into $K$ action state prompt embedding features, represented as $\textbf{{p}}_{s} = [\textbf{p}_{{s}_1}, \ldots, \textbf{p}_{{s}_{K}}]^\top \in \mathbb{R}^{K \times d}$, where $d$ is the hidden dimension of CLIP. 
	
	\noindent \textbf{Prompt Adapter.} We feed different types of action-aware prompts $\textbf{p}_{t}$ and $\textbf{p}_{s}$ into a prompt adapter, which consists of two Multilayer Perceptron (MLP) layers to generate prompt embedding features. Between these layers, a ReLU activation and layer-normalization are applied to generate the action-aware visual prompts:
	
	\begin{equation}
	\begin{split}
		\textbf{p}^{t}_{0} &= \text{ReLU}\left(\text{LN}\left(\textbf{p}_{t}\right) \cdot \textbf{W}_{\text{down}}\right) \cdot \textbf{W}_{\text{up}}, \\
		\textbf{p}^{s}_{0} &= \text{ReLU}\left(\text{LN}\left(\textbf{p}_{s}\right) \cdot \textbf{W}_{\text{down}}\right) \cdot \textbf{W}_{\text{up}},
	\end{split}
	\end{equation}
	where the feature dimension is first scaled from \( d \) to \( t \) (\( t \ll d \)) by the learnable matrix \( \textbf{W}_{\text{down}} \in \mathbb{R}^{d \times t} \) and then expanded back to \( d \) by \( \textbf{W}_{\text{up}} \in \mathbb{R}^{t \times d} \). 
	
	\noindent \textbf{Action-aware Adaptive Interaction Module.}
	To reduce the disturbance caused by irrelevant or noise information in the prompts, we construct an action-aware adaptive interaction module to adaptively tune the visual features by focusing on key information relevant to actions, thereby allowing the prompted image features to better align with the corresponding textual features, as shown in the right part of Figure~\ref{pipeline}.
	
	Specifically, we first divide the input image $\mathrm{\textbf{X}} \in \mathbb{R}^{H \times W \times C}$ into a sequence of $M$ image patches, denoted as $\mathrm{\textbf{X}} = \{\textbf{x}_p^1, \textbf{x}_p^2, \ldots, \textbf{x}_p^M\}$, $\mathrm{\textbf{X}} \in \mathbb{R}^{M \times d_{v}}$. 
	Subsequently, each patch is transformed into an embedding vector and augmented with a learnable position embedding. The pre-processed image patch embedding for the transformer input is represented as $\textbf{E}_0 = [\textbf{e}_0^1, \textbf{e}_0^2, \ldots, \textbf{e}_0^M]$, where $\textbf{e}_0^i \in \mathbb{R}^{C_e}$ is the patch embedding corresponding to position $i$, with $C_e$ being the number of channels for each embedding. We take pre-processed image patch embedding $\textbf{E}_{0}$, action triplet prompt embedding features $\textbf{{p}}^{t}_{0}$ and action state prompt embedding features $\textbf{p}^{s}_{0}$ as query, key, and value, formulated as
	\begin{equation} 
	\begin{split}
		\tilde{\textbf{p}}^{t}_{0}
		= \text{Cross-Attn}(\textbf{p}^{t}_{0},\textbf{E}_0), 
		\tilde{\textbf{p}}^{s}_{0}
		= \text{Cross-Attn}(\textbf{p}^{s}_{0},\textbf{E}_0), \\
		\textbf{{V}}^{t}_{cs}, \textbf{{A}}^{t}_{cs} = \text{Self-Attn}(\tilde{\textbf{p}}^{t}_{0}), 
		\textbf{{V}}^{s}_{cs}, \textbf{{A}}^{s}_{cs} = \text{Self-Attn}(\tilde{\textbf{p}}^{s}_{0}).
	\end{split}
	\end{equation}
	where $\textbf{{A}}^{t}_{cs}$ and $\textbf{{A}}^{s}_{cs}$ are the attention matrix, $\text{Cross-Attn}(\cdot)$ and $\text{Self-Attn}(\cdot)$ denote the cross-attention and self-attention operation, respectively. In this manner, we make the original patch feature attend to action triplet prompts and action state prompts, resulting in augmented image features. These features are then transformed by MLP and a re-weighting strategy. The whole process with attention is formulated as 
	\begin{equation}
	\begin{split}
		\textbf{V} = \text{MLP}([\textbf{{V}}^{t}_{cs}, \textbf{{V}}^{s}_{cs}]), 
		\tilde{\textbf{V}}
		&= (\lambda \textbf{{A}}^{t}_{cs} + (1 - \lambda) \textbf{{A}}^{s}_{cs}) \textbf{V},
	\end{split}
	\end{equation}
	where $[\cdot]$ indicates the concatenation operation, 
	$\tilde{\textbf{V}}
	= \{\textbf{v}_1, \textbf{v}_2, \ldots, \textbf{v}_{L}\} \in \mathbb{R}^{L \times d}$ denotes the action-enhanced visual features, and $\lambda$ is a hyper-parameter with a value of 0.7. Afterward, we randomly initialize learnable visual vector, defined as $\textbf{P}_{v} = [\textbf{P}]_1 [\textbf{P}]_2 \ldots [\textbf{P}]_N, \text{ where } [\textbf{P}]_n (n \in \{1, \ldots, N\}) $ keeps the same dimension as the image patch embedding. We concatenate the learnable visual prompts and action-enhanced visual features interlaced, resulting in the final visual features $\textbf{{z}}'_{l}= [\textbf{v}_1, \textbf{v}_2, \ldots, \textbf{v}_{L}, [\textbf{P}]_1 [\textbf{P}]_2 \ldots [\textbf{P}]_N].$
	Subsequently, we feed them into the rest $L$-layer transformer layers of visual encoder $\theta$ for generating the action-aware visual embedding: 
	\begin{equation}
	\textbf{{z}}'_{i} = \theta_{i}(\textbf{{z}}'_{i-1}), \; i \in [l + 1, L].
	\end{equation}
	The output of the last layer $\textbf{z}^{'}_{l}$ is treated as the prompted image features $\textbf{z}^{img}$ used for optimization with contrastive loss and triplet loss in Eq.~\ref{loss_i2t} and Eq.~\ref{loss2}.
	
	\subsection{Training}
	\label{sec:training}
	We introduce a two-stage training procedure to learn an action-aware vision-language model.
	
	\noindent \textbf{The first training stage.} In this stage, the objective is to update newly added parameters of action-aware prompts, triplet encoder, and action-aware adaptive interaction module while keeping the original weights of CLIP frozen. Based on the prompted image features and textual features, the contrastive loss is defined by
	\begin{align}
	\mathcal{L}_{\text{stage1}} &= \mathcal{L}_{\text{i2t}} + \mathcal{L}_{\text{t2i}}, \notag \\
	\mathcal{L}_{\text{i2t}}(i) &= -\log \frac{\exp(\text{sim}(\textbf{z}^{img}_{i}, \textbf{z}^{text}_{i}) / \tau)}{\sum_{j=1}^{N} \exp(\text{sim}(\textbf{z}^{img}_{i}, \textbf{z}^{text}_{j}) / \tau)}, \label{loss_i2t} \\
	\mathcal{L}_{\text{t2i}}(i) &= -\log \frac{\exp(\text{sim}(\textbf{z}^{img}_{i}, \textbf{z}^{text}_{i}) / \tau)}{\sum_{j=1}^{N} \exp(\text{sim}(\textbf{z}^{img}_{j}, \textbf{z}^{text}_{i}) / \tau)}, \notag \label{loss_t2i}
	\end{align}
	\label{loss1}
	where $\textbf{z}^{text}$ denotes the final text embeddings encoded by a frozen text encoder given an input caption $T$, $\cos(\cdot)$ is the cosine similarity between the inputs, and $\tau$ is a learnable temperature parameter. The terms $\mathcal{L}_{\text{i2t}}$ and $\mathcal{L}_{\text{t2i}}$ refer to image-to-text and text-to-image contrastive losses, respectively. By minimizing $\mathcal{L}_{\text{i2t}}$ and 
	$\mathcal{L}_{\text{t2i}}$, the action-aware prompts will be updated during training through gradient backpropagation.
	
	\begin{table*}[htbp]
	\centering
	\setlength{\tabcolsep}{3pt} 
	\scriptsize 
	\begin{tabular}{l|l|ccc|ccc|c|ccc|ccc|ccccc}
		\hline
		\multirow{3}{*}{Backbone} & \multirow{3}{*}{Method} & \multicolumn{6}{c|}{COCO (5K test images)} & \multirow{3}{*}{Rsum} & \multicolumn{6}{c|}{Flickr30K (1K test images)} & \multirow{3}{*}{Rsum} \\
		\cline{3-8} \cline{10-15}
		& & \multicolumn{3}{c|}{Image-to-Text} & \multicolumn{3}{c|}{Text-to-Image} & & \multicolumn{3}{c|}{Image-to-Text} & \multicolumn{3}{c|}{Text-to-Image} & \\
		\cline{3-5} \cline{6-8} \cline{10-12} \cline{13-15}
		& & R@1 & R@5 & R@10 & R@1 & R@5 & R@10 & & R@1 & R@5 & R@10 & R@1 & R@5 & R@10 & \\
		\hline
		\multirow{5}{*}{ViT-B/32}
		& CLIP~\cite{radford2021learning} & 50.2 & 74.6 & 83.6 & 30.4 & 56.0 & 66.8 & 361.6 & 79.0 & 94.3 & \underline{98.2} & 58.0 & 82.9 & 89.9 & 502.3 \\
		& MGCA~\cite{wang2022multi} & {54.5} & \underline{78.6} & \underline{86.8} & 37.7 & 63.7 & 74.0 & \underline{395.3} & 81.5 & 93.9 & 96.8 & 64.4 & 86.5 & 92.0 & \underline{515.1} \\
		& PyramidCLIP~\cite{gao2022pyramidclip} & 52.8 & 78.1 & - & 38.8 & 64.9 & - & - & \underline{84.2} & \underline{96.4} & - & \underline{69.1} & \underline{89.8} & - & - \\
		& CLIPDualDIS~\cite{DBLP:conf/emnlp/WangJ023} & - & - & - & 37.9 & 63.4 & 73.4 & - & - & - & - & 59.0 & 83.4 & 90.1 & - \\
		&OpenCLIP~\cite{cherti2023reproducible} &- &- &- &\underline{39.4} &\underline{65.5} &\underline{75.7} &- &- &- &-  &63.9 &87.3 &\underline{93.2} &- \\
        &SaCo~\cite{wu2024saco} &\underline{54.6} &77.6  &- &36.1 &62.0 &- &- &81.7 &95.5 &- &64.9 &88.1 &- &-\\
		& Ours & \textbf{55.5} & \textbf{79.2} & \textbf{87.8} & \textbf{39.9} & \textbf{66.0} & \textbf{76.3} & \textbf{404.7} & \textbf{85.1} & \textbf{97.4} & \textbf{98.3} & \textbf{69.9} & \textbf{90.1} & \textbf{94.8} & \textbf{535.6} \\
		\hline
		\multirow{5}{*}{ViT-B/16}
		& CLIP~\cite{radford2021learning} & 52.4 & 76.9 & 84.8 & 33.1 & 58.5 & 69.2 & 374.9 & 81.2 & 96.4 & 98.5 & 62.2 & 85.7 & 91.8 & 515.8 \\
		& MGCA~\cite{wang2022multi} & 57.6 & 80.5 & {87.8} & 39.8 & 65.7 & {75.3} & 406.7 & 82.2 & 96.1 & 98.1 & 67.7 & 88.5 & 93.2 & 525.8 \\
		& PyramidCLIP~\cite{gao2022pyramidclip} & 55.7 & \underline{80.8} & - & \underline{42.6} & \underline{68.6} & - & - & {85.6} & \underline{97.7} & - & \underline{74.5} & \underline{92.9} & - & - \\
		&EVA-02-CLIP~\cite{sun2023eva} &\textbf{58.7} &80.7 &\underline{88.2} &42.2 &66.9 &\underline{76.3} &\underline{413.1} &\underline{85.7} &96.7 &\underline{98.9} &71.2 &91.0 &\underline{94.7} &\underline{538.2} \\
		& FineCLIP~\cite{jingfineclip} & {54.5} &78.6 &85.8 &40.2 &66.5 &76.1 &401.7  & 82.5 &96.4 &98.6 &67.9 &89.1 &{94.1} &528.6 \\
        &SaCo~\cite{wu2024saco} &57.8 &80.0 &- &39.8 &64.7 &- &- &85.5 &96.5 &- &69.1 &90.1 &- &-\\
		& Ours & \underline{58.4} & \textbf{81.8} & \textbf{89.1} & \textbf{43.2} & \textbf{69.5} & \textbf{79.0} & \textbf{421.1} & \textbf{88.1} & \textbf{98.0} & \textbf{99.2} & \textbf{74.7} & \textbf{93.1} & \textbf{95.8} & \textbf{548.9} \\
		\hline
		\multirow{6}{*}{ViT-L/14-336}
		& CLIP~\cite{radford2021learning} & 57.3 & 80.6 & 87.8 & 37.9 & 63.4 & 73.4 & 394.2 & 86.6 & 98.0 & 99.1 & 67.1 & 88.9 & 93.2 & 532.9 \\
		& MGCA~\cite{wang2022multi} & 59.7 & 83.2 & 89.7 & 44.3 & 69.6 & 78.8 & 425.3 & 86.9 & 97.3 & 98.6 & 74.4 & 91.7 & 95.4 & 544.3 \\
		& FILIP~\cite{DBLP:conf/iclr/YaoHHLNXLLJX22} & 61.3 & 84.3 & 90.4 & 45.9 & 70.6 & 79.3 & 431.8 & \underline{89.8} & \underline{99.2} & \textbf{99.8} & {75.0} & 93.4 & {96.3} & {553.5} \\
		& REACT~\cite{liu2023learning} & \underline{63.3} & 85.1 & - & \underline{47.5} & 72.0 & - & - & 90.4 & 99.1 & - & \underline{76.5} & {93.7} & - & - \\
		&MetaCLIP~\cite{wang2024diffusion} &\textbf{65.5} &\underline{85.2} &\underline{91.1} &\textbf{48.2} &\underline{72.3} &\underline{81.1} &\textbf{443.4} &89.5 &98.8 &\underline{99.7} &\textbf{76.8} &\underline{93.9} &\underline{96.6} &\underline{555.3}\\
		& Ours & 62.5 & \textbf{85.7} & \textbf{91.6} & 44.1 & \textbf{72.9} & \textbf{81.7} & \underline{438.5} & \textbf{91.5} & \textbf{99.5} & \textbf{99.8} & 74.0 & \textbf{94.4} & \textbf{97.6} & \textbf{556.7} \\
		\bottomrule
	\end{tabular}
	\caption{Comparative results (without pre-training on external training datasets) on the COCO 5K test set and Flickr30K test set. The best results and the second best results are  marked by \textbf{bold} and \underline{underline}, respectively.}
	\label{combined}
    \vspace{-5pt}
	\end{table*}
	
	\noindent \textbf{The second training stage.} In this stage, only the parameters of the action-aware adaptive interaction module in visual encoder are optimized while keeping the action-aware prompts and prompt adapter frozen. To boost the final performance, we employ the triplet loss in our second training stage for optimization, given by
	\begin{equation}
	\mathcal{L}_{\text{stage2}} = \max(d_p - d_n + \alpha, 0),
	\label{loss2}
	\end{equation}
	where $d_p$ and $d_n$ represent the feature distances of the positive pair and negative pair, respectively, and $\alpha$ is the margin of $\mathcal{L}_{\text{tri}}$.
	
	The whole training process utilizes the action-aware prompts to mine and store the hidden states of the pre-trained image encoder, enabling CLIP to retain its advantages. The first stage focuses on learning effective prompts to fully harness the capabilities of CLIP, while the second stage refines multi-modal interactions within a more fine-grained embedding space.

    \subsection{Inference}
    First of all, the original CLIP model is used to pre-select the top-k  (\textit{i.e.}, k=20) samples, after which our method is applied to refine features and re-rank results, enabling fast retrieval.
    Specifically, for image-to-text retrieval, we first extract text features for pre-selected top-k gallery texts and enrich their semantics using LLM-generated action triplets and state descriptions. Then we extract the image feature enhanced by the enriched texts. Finally, we retrieve the target text by calculating the similarities between the enhanced image feature and all enhanced text features. A similar procedure is adopted for text-to-image retrieval.
\section{Experiments}
	\subsection{Datasets}
	To evaluate the effectiveness of the proposed method, we conduct extensive experiments using two datasets: Flickr30K~\cite{young2014image}  and COCO~\cite{chen2015microsoft}. More details about the statistics of these datasets are reported in the supplementary material.
	
	\subsection{Evaluation Metrics}
	To evaluate the image-text matching performance, we measure the proportion of queries that match the correct item within the top-k results, denoted as R@K, where K takes values of 1, 5, and 10. The sum of all R@K values is calculated to evaluate the overall matching performance, denoted as Rsum. More details about implementation can be found in the supplementary material.
	
	\subsection{Comparison with State-of-the-Art Methods}
	To verify the effectiveness of the proposed method, we compare it with several state-of-the-art methods on the Flickr30K and COCO datasets. For fair comparisons, all these methods are CLIP-based without pre-training. 
	
	The comparison results on the two datasets are shown in Table~\ref{combined}. 
	We have the following observations: 1) Our method achieves significant performance improvements over the  pre-trained CLIP models in  different ViT architectures. This clearly indicates that enriching the fine-grained action semantic information of visual modality helps image-text alignment. 2) Compared to other CLIP-based methods, Our method achieves competitive results on both datasets  in terms of most evaluation metrics, demonstrating the effectiveness of incorporating action knowledge into prompt tuning. 
	
	It is worth noting that our method achieves significant performance improvements using the ViT-B backbone compared to those methods based on the ViT-L backbone. The possible reason is that the ViT-B architecture has fewer layers, smaller hidden dimensions, and fewer attention heads, and is inherently less capable of capturing and processing fine-grained image details. Our method can address this limitation by injecting action knowledge into CLIP through action-aware multi-modal prompt tuning, enabling the model to achieve a more fine-grained action perception ability. 
	
	\subsection{Ablation Study}
	We perform in-depth ablation studies to evaluate each component of our method on the COCO dataset, using pre-trained CLIP with a ViT-B/16 backbone as the baseline. 
	We report the results on the Flickr30K dataset in the supplementary material.
	
	\begin{table}[h!]
		\centering
        \vspace{-10pt}
		\setlength{\tabcolsep}{3pt} 
		\renewcommand{\arraystretch}{1.0} 
		\scriptsize 
		\begin{tabular}{cccc|ccc|ccc}
			\hline     
			\multirow{2}{*}{\makecell{Action \\ Triplet}} & \multirow{2}{*}{\makecell{Hand-Craft \\ Triplet}} & \multirow{2}{*}{\makecell{Action \\ State}} & \multirow{2}{*}{Vis} & \multicolumn{3}{c|}{Image-to-Text} & \multicolumn{3}{c}{Text-to-Image} \\
			\cline{5-10}
			& & & & R@1 & R@5 & R@10 & R@1 & R@5 & R@10 \\
			\hline
			\ding{51} & & & & 56.2 & 81.0 & 88.4 & 42.4 & 68.9 & 78.3 \\
			& \ding{51} & & & 55.5 & 80.7 & 88.1 & 41.5 & 68.1 & 77.9 \\
			& & \ding{51} & & 53.7 & 80.9 & 88.5 & 42.3 & 69.1 & 78.9 \\
			\ding{51} & & \ding{51} & & 58.1 & 81.8 & 88.5 & 42.9 & 69.3 & 78.7 \\
			\ding{51} & & \ding{51} & \ding{51} & \textbf{58.4} & \textbf{81.8} & \textbf{89.1} & \textbf{43.2} & \textbf{69.5} & \textbf{79.0} \\
			\hline
		\end{tabular}
		\caption{Ablation studies of prompts on the COCO 5K test set. “Hand-Craft Triplet” denotes replacing the action triplet prompt with a hand-crafted prompt template, \textit{i.e.}, “A photo capturing a [subject] performing [action] in relation to [object].”, and “Vis” denotes visual prompting.}
		\label{ablation2}
        \vspace{-10pt}
	\end{table}

	\noindent \textbf{Effectiveness of action-aware multi-modal prompting.}
	To evaluate the effectiveness of action-aware multi-modal prompting, we design several variants of our method for comparison. 
	From the results shown in Table~\ref{ablation2}, it is obvious that the proposed action-aware multi-modal prompting outperforms other variants. Specifically, removing either the action triplet prompt or the action state prompt leads to a drop in matching performance. The possible reason is that the action triplet prompt is designed to explicitly model action relations between entities, while the action state prompt is designed to provide additional fine-grained information from action states to mitigate ambiguity in action relationships. The combination of these two types of action prompts can further improve the performance. In addition, removing the visual prompt weakens the model's visual perception ability, thereby hurting the performance.
	
	\noindent\textbf{Effectiveness of action knowledge.} To evaluate the effectiveness of incorporating action knowledge in prompting the pre-trained CLIP, we replace the action-aware multi-modal prompts with learnable visual prompts, which simplifies the framework to a standard visual prompt tuning for the vision encoder. As shown in Table~\ref{ablation1}, we observe that the performance drops significantly. This is expected, as incorporating action knowledge enhances the fine-grained visual perception ability of the pre-trained model.

    \noindent \textbf{Effective of attribute knowledge.} To evaluate the impact of attribute knowledge in promoting pre-trained CLIP, we remove attribute knowledge from action triplets. As shown in Table~\ref{ablation1}, the results indicate a slight drop in performance, as attribute knowledge provides more fine-grained details about the states of object attributes, enriching the overall semantic understanding of actions.
	
	\begin{table}[h!]
    \vspace{-8pt}
		\centering
		\setlength{\tabcolsep}{2.1pt} 
		\renewcommand{\arraystretch}{1.0} 
		\small 
		\begin{tabular}{l|ccc|cccc}
			\hline     
			\multirow{2}{*}{Method} & \multicolumn{3}{c|}{Image-to-Text} & \multicolumn{3}{c}{Text-to-Image} \\
			\cline{2-7}
			& R@1 & R@5 & R@10 & R@1 & R@5 & R@10 \\
			\hline
			Baseline & 52.4 &76.9 &84.8 &33.1 &58.5 &69.2 \\ 
			w/o action knowledge &55.3 &78.8 &86.8 &41.5 &67.3 &76.9 \\
            w/o attribute knowledge &58.2	&81.6	&88.6	&43.1	&68.6	&78.2\\
			replace AIM with CAT &57.4 &79.9 &87.6 &41.9 &67.8 &77.4 \\
			Ours &\textbf{58.4} &\textbf{81.8} &\textbf{89.1} &\textbf{43.2} &\textbf{69.5} &\textbf{79.0} \\
			\hline
		\end{tabular}
		\caption{Ablation analysis of different components on the COCO 5K test set.}
        \vspace{-18pt}
		\label{ablation1}
	\end{table}

    \begin{table}[h!]
    \centering
    \setlength{\tabcolsep}{2.1pt} 
    \renewcommand{\arraystretch}{1.0} 
    \small 
    \begin{tabular}{l|c|ccc|ccc}
        \hline
        \multirow{2}{*}{LLMs} & \multirow{2}{*}{version} & \multicolumn{3}{c|}{Image-to-Text} & \multicolumn{3}{c}{Text-to-Image} \\
        \cline{3-8}
        & & R@1 & R@5 & R@10 & R@1 & R@5 & R@10 \\
        \hline
        Llama-2 & Vicuna-7B & 58.4 & 81.4 & 89.0 & 43.0 & 69.2 & 78.8 \\
        Llama-2 & Vicuna-13B & \textbf{58.5} & 81.7 & \textbf{89.1} & 43.2 & 69.4 & \textbf{79.1} \\
        GPT-3.5 & GPT-3.5-Turbo & 58.4 & \textbf{81.8} & \textbf{89.1} & \textbf{43.2} & \textbf{69.5} & {79.0} \\
        \hline
    \end{tabular}
    \caption{Ablation study of different LLMs.}
    \label{retrieval_comparison}
    \vspace{-8pt}
\end{table}
	
	\noindent \textbf{Effectiveness of action-aware adaptive interaction.} To evaluate the effectiveness of action-aware adaptive interaction, we replace the action-aware adaptive interaction module (denoted as “AIM”) with a concatenation operation (denoted as “CAT”). Specifically, we directly concatenate action triplet prompts, action state prompts, and learnable visual prompts to generate multi-modal prompts for prompt tuning. The results are reported in Table~\ref{ablation1}, which shows that using concatenation operation degrades performance. This demonstrates the vital importance of AIM as it can focus on key action information and aggregate it to generate attentive features, thereby reducing the disturbance caused by the irrelevant or noisy information retained in the action triplet prompts and action state prompts.
	
\begin{table}[h!]
    \centering
     \vspace{-10pt}
    \setlength{\tabcolsep}{2.0pt} 
    \renewcommand{\arraystretch}{1.2} 
    \small 
    \begin{tabular}{l|ccc|ccc}
        \hline
        \multirow{2}{*}{Method} & \multicolumn{3}{c|}{Image-to-Text} & \multicolumn{3}{c}{Text-to-Image} \\
        \cline{2-4} \cline{5-7}
         & R@1 & R@5 & R@10 & R@1 & R@5 & R@10 \\
        \hline
        Baseline & 52.4 &76.9 &84.8 &33.1 &58.5 &69.2\\
        combined training &56.5 &80.2 &87.9 &42.1 &68.0 &77.6\\
        one-stage training &56.8 &80.5 &87.8 &41.9 &67.8 &77.8\\
        two-stage training (Ours) &\textbf{58.4} &\textbf{81.8} &\textbf{89.1} &\textbf{43.2} &\textbf{69.5} &\textbf{79.0}  \\
        \hline
    \end{tabular}
    \caption{Ablation study of different training.}
    \vspace{-8pt}
    \label{ablation3}
\end{table}

    \noindent \textbf{Effectiveness of different LLMs.}
    To explore the effect of using different types and sizes of LLMs (\textit{i.e.}, Llama-2-7B, Llama-2-13B, GPT-3.5), we conduct an ablation study on the COCO dataset. As shown in Table~\ref{retrieval_comparison}, the results indicate that using different types and sizes of LLMs has little impact on performance, demonstrating that even smaller open-source models can achieve comparable results while reducing costs and enhancing reproducibility of our method.

    \noindent \textbf{Effect of two-stage training.} 
To evaluate the effectiveness of the proposed two-stage training strategy, we design two different training strategies for comparison:

\begin{itemize}
    \item \textbf{One-stage training}: we only use \(\mathcal{L}_{\text{stage1}}\) to jointly train the triplet encoder and action-aware adaptive interaction simultaneously.
    \item \textbf{Combined Training}: we optimize the model by combining \(\mathcal{L}_{\text{stage1}}\) and \(\mathcal{L}_{\text{stage2}}\) into a unified training process rather than separating the training into two stages.
\end{itemize}

{\color{black}{As shown in Table~\ref{ablation3}, the results demonstrate that both one-stage training and combination training are less effective. In the early stage of training, the learnable visual tokens cannot effectively describe the image, which negatively impacts the optimization of the image encoder. Additionally, combination training struggles to achieve desirable results, as the simultaneous optimization of \(\mathcal{L}_{\text{stage1}}\) and \(\mathcal{L}_{\text{stage2}}\) leads to interference between objectives, which hinders effective feature learning. In contrast, our proposed two-stage training strategy achieves better performance. During the first stage, the model focus more on the comparison between positive samples and negative samples without explicit fixed distance constraints, which can accelerate training. During the second stage, the model establishes stable intra-class and inter-class relationships, further enlarging feature distances between categories and enhancing sensitivity to subtle feature differences. }}
	
	\begin{figure}
		\centering
		\includegraphics[width=1.0\linewidth, keepaspectratio]{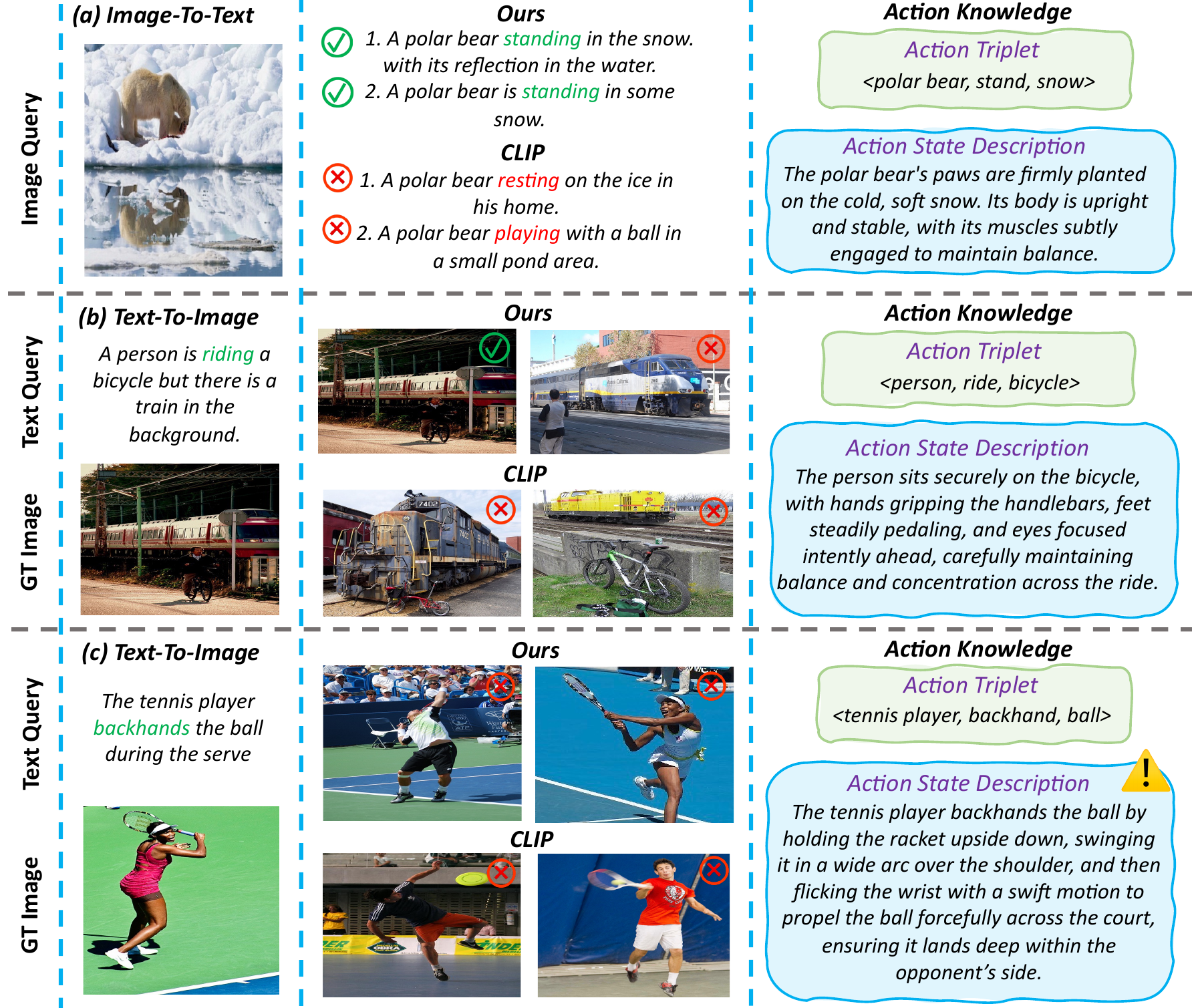}
		\caption{The qualitative results on the COCO 5K test set, where correct matches are marked in {\protect\greencheck} and mismatches in {\protect\redcross}.}
		\label{ablation5}
        \vspace{-8pt}
	\end{figure}
	\subsection{Qualitative Results}
	We have presented the comparison results between our method and CLIP in Figure~\ref{ablation5}. As shown in Figures~\ref{ablation5} (a) and (b), given a text query or an image query, with the help of action knowledge from LLM, our method retrieves top-2 ranked results matching the query. In Figure~\ref{ablation5} (a), the action knowledge of “standing” is obviously different from actions such as “resting” or “playing”. In Figure~\ref{ablation5} (b), with the precise description of “riding”, our method effectively filters out images of bicycles or trains.
	However, as shown in Figure~\ref{ablation5} (c), the action knowledge generated by the LLM for “\textless tennis player, backhand, ball\textgreater” can not accurately capture the action state of “backhand ball”, resulting in incorrect retrieval.
	A more advanced LLM may be able to solve this problem.
	More visualization results and analyses of retrieval results can be found in the supplementary material.
	\section{Conclusion}
	We have presented a simple yet effective LLM-enhanced action-aware multi-modal prompt-tuning method that integrates external action knowledge to prompt the pre-trained vision-language model. 
	We propose an action triplet prompt and an action state prompt that use the knowledge of LLM to help CLIP concentrate on detailed action-related visual cues. 
	We also design an action-aware adaptive interaction module to aggregate key information from different prompts, enhancing the action-aware visual representations.  
	Extensive experiments on the COCO and Flickr30K datasets demonstrate the effectiveness of our method. In the future, we plan to apply our method to tasks such as video-text matching and action recognition, where LLMs can assist in perceiving detailed actions.
	Furthermore, incorporating LLMs into the training process of CLIP-like methods may be a potential direction.

    \section{Acknowledgments}
    This work was supported in part by the grants from the Shenzhen Science and Technology Program under Grant No.~JCYJ20241202130548062, the Natural Science Foundation of Shenzhen under Grant No.~JCYJ20230807142703006, and the Key Research Platforms and Projects of the Guangdong Provincial Department of Education under Grant No.~2023ZDZX1034.

{
    \small
    \bibliographystyle{ieeenat_fullname}
    \bibliography{main}
}
\newpage
\appendix
\section{Datasets Details}
\noindent \textbf{Flickr30K}~\cite{young2014image} dataset contains 31,000 images collected from the Flickr website. These images mostly depict humans performing various activities. Each image is described by five different sentences, and there are 155,000 sentences. Following the settings~\cite{wang2020consensus,chen2021learning},
this dataset is split into 29,783 training images, 1,000 validation images and 1,000 testing images.

\noindent \textbf{COCO}~\cite{chen2015microsoft} dataset is a challenging large-scale dataset containing 123,287 images. These images are collected by searching 80 object categories and 40 scene types from the Flickr website. Each image is paired with five sentences, resulting in a total of 616,435 sentences. We follow the dataset split in works~\cite{wang2020consensus,chen2021learning}, namely, 113, 287 images for training, 5,000 images for validation, and 5,000 images for testing.

\section{Implementation Details}
In all experiments, we use GPT-3.5 as the LLM and CLIP 
as the pre-trained model. We use three popular pre-trained backbones of CLIP: ViT-B/32, ViT-B/16, and ViT-L/14-336. During training, we use SGD optimization with an initial learning rate of 1e-5, a maximum of 4 epochs, and a batch size of 128.

\section{More Ablation Studies}
\setlength{\intextsep}{0pt}  
\setlength{\columnsep}{10pt}  

\noindent \textbf{Effect of action-aware multi-modal prompting.} To evaluate the effectiveness of action-aware multi-modal prompting, we design several variants of our method for comparison. From the results in Table~\ref{ablation_flickr}, we observe that our method outperforms the other variants on the Flickr30K dataset. These results highlight the importance of both the action triplet prompt and the action state prompt in enhancing matching performance, as each captures fine-grained action semantics. Moreover, our method incorporates the visual prompt to enhance the model's visual perception.

\begin{table}[h!]
    \centering
    \vspace{0.3cm}
    \setlength{\tabcolsep}{3pt} 
    \renewcommand{\arraystretch}{1.0} 
    \scriptsize 
    \begin{tabular}{c|c|c|c|ccc|ccc}
        \hline     
        \multirow{2}{*}{\makecell{Action \\ Triplet}} & \multirow{2}{*}{\makecell{Hand-Craft \\ Triplet}} & \multirow{2}{*}{\makecell{Action \\ State}} & \multirow{2}{*}{\makecell{Vis}} & \multicolumn{3}{c|}{Image-to-Text} & \multicolumn{3}{c}{Text-to-Image} \\
        \cline{5-10}
        & & & & R@1 & R@5 & R@10 & R@1 & R@5 & R@10 \\
        \hline
        \ding{51} & & & & 86.1 & 97.6 & 98.6 & 73.4 & 92.7 & 95.4 \\
        & \ding{51} & & & 85.3 & 97.2 & 98.2 & 72.7 & 92.5 & 95.1 \\
        & & \ding{51} & & 86.7 & 97.2 & 98.6 & 74.0 & 92.9 & 95.8 \\
        \ding{51} & & \ding{51} & & 87.5 & 97.8 & 98.8 & 74.3 & 92.9 & 95.6 \\
        \ding{51} & & \ding{51} & \ding{51} & \textbf{88.1} & \textbf{98.0} & \textbf{99.2} & \textbf{74.7} & \textbf{93.1} & \textbf{95.8} \\
        \hline
    \end{tabular}
    \caption{Ablation study over types of prompts on Flickr30K 1K test set.}
    \label{ablation_flickr}
\end{table}

\noindent\textbf{Effect of action knowledge.} To evaluate the effectiveness of incorporating action knowledge in prompting the pre-trained CLIP, we replace the action-aware multi-modal prompts with learnable visual prompts, denoted as ``w/o action knowledge". As shown in Table~\ref{ablation_flickr1}, the performance of our method with only visual prompt tuning drops significantly, indicating that incorporating prompts sourced from action-related external knowledge is crucial for enhancing the fine-grained visual perception ability of the pre-trained model.

\begin{table}[h!]
    \centering
    \vspace{0.5cm}
    \setlength{\tabcolsep}{2.1pt} 
    \renewcommand{\arraystretch}{1.2} 
    \small 
    \begin{tabular}{l|ccc|ccc}
        \hline
        \multirow{2}{*}{Method} & \multicolumn{3}{c|}{Image-to-Text} & \multicolumn{3}{c}{Text-to-Image} \\
        \cline{2-7}
         & R@1 & R@5 & R@10 & R@1 & R@5 & R@10 \\
        \hline
        Baseline & 81.2 &96.4 &98.5 &62.2 &85.7 &91.8 \\
        w/o action knowledge &86.8 &97.2 &98.9 &69.7 &90.3 &93.1 \\
        replace AIM with CAT &87.2 &97.9 &99.1 &72.9 &91.2 &95.6 \\
        replace PA with FC &87.9 &98.0 &99.1 &73.2 &92.0 &95.9 \\
        Ours &\textbf{88.1} &\textbf{98.0} &\textbf{99.2} &\textbf{74.7} &\textbf{93.1} &\textbf{95.8} \\
        \hline
    \end{tabular}
    \vspace{0.3cm}
    \caption{Ablation analysis of different components on Flickr30K 1K test set.}
    \label{ablation_flickr1}
\end{table}

\noindent \textbf{Effect of action-aware adaptive interaction.}
In Table~\ref{ablation_flickr1}, we evaluate the effect of replacing the action-aware adaptive interaction module (denoted as “AIM”) with a concatenation operation (denoted as “CAT”) on the Flickr30K dataset. We observe a decline in performance, indicating that AIM helps the model mitigate the disturbance caused by irrelevant or noisy information retained in prompts.

\noindent \textbf{Effect of prompt adapter.} To evaluate the effectiveness of the prompt adapter (denoted as “PA”) in adapting the prompted knowledge, we replace it with a simple fully connected layer (denoted as “FC”). As shown in Table~\ref{ablation_flickr1}, the prompt adapter performs best. This is reasonable since prompt tuning and adapter play different roles in improving performance, in which the prompt adapter further bridges the feature gap between the pre-trained model and the downstream task.  

\begin{figure}[htbp]
\vspace{0.3cm}
  \centering
  \begin{subfigure}{0.22\textwidth}
    \centering
    \includegraphics[width=\textwidth]{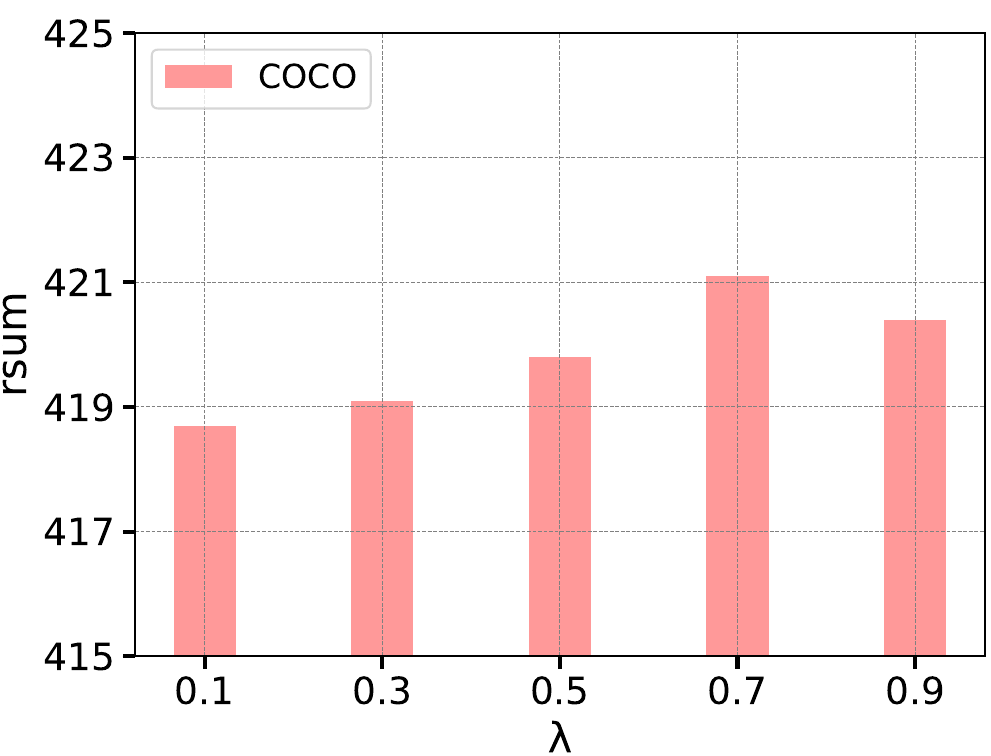}
    \caption{COCO}
  \end{subfigure}
\hspace{0.4cm}
  \begin{subfigure}{0.22\textwidth}
    \centering
    \includegraphics[width=\textwidth]{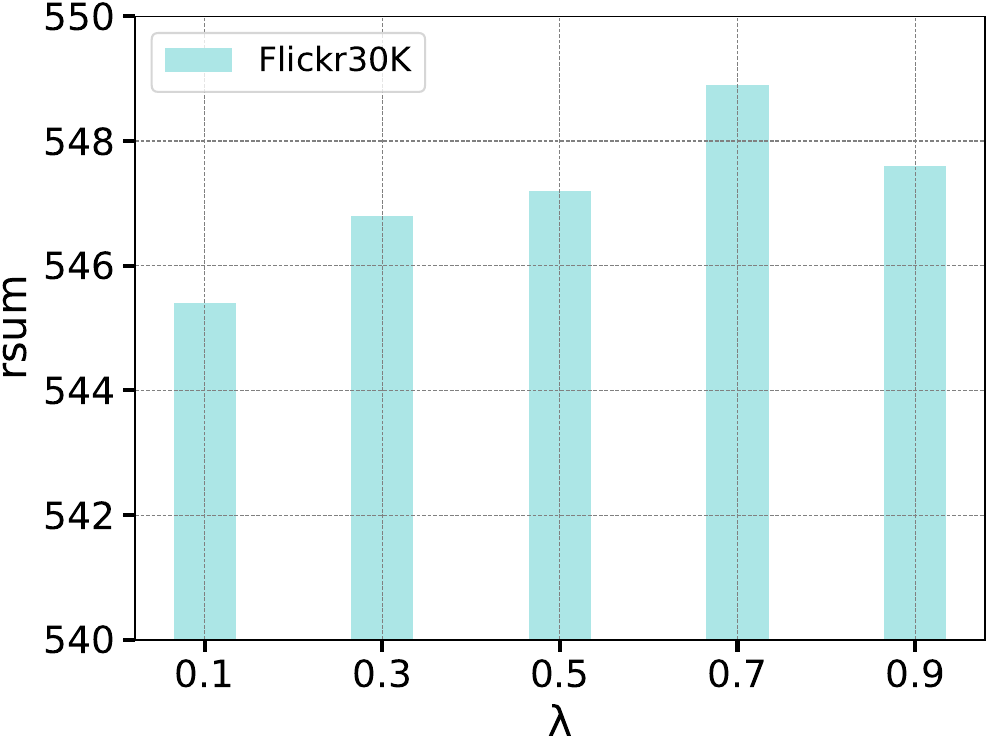}
    \caption{Flickr30K}
  \end{subfigure}
  \caption{Comparison results of different trade-off hyper-parameters on COCO 5K test set and Flickr30K 1K test set.}
  \label{ablation6}
\end{figure}

\noindent \textbf{Effect of Hyper-parameter $\lambda$.} We evaluate the impact of the trade-off hyper-parameter $\lambda$ in Eq.~9 by varying its values among 0.1, 0.3, 0.5, 0.7, and 0.9. As depicted in Figure.~\ref{ablation6}, our method achieves the best performance when $\lambda$ is set to 0.7 on both datasets.

\begin{figure*}[h!]
    \centering
    \includegraphics[width=0.8\textwidth]{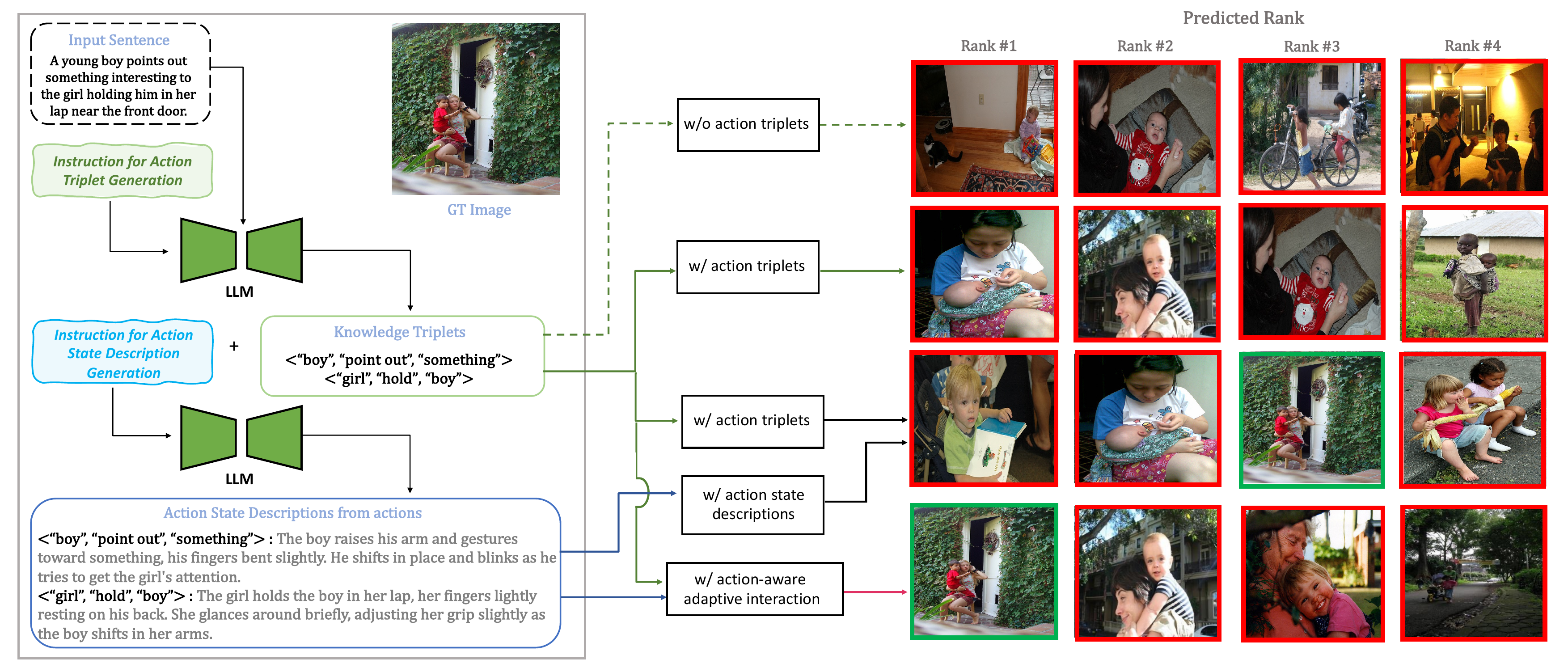}
    \caption{An example from the ablation study in text-to-image retrieval, comparing various variants (\textit{i.e.}, "w/o action triplets," "w/ action triplets," "w/ action triplets and action state descriptions," and "w/ action-aware adaptive interaction").}
    \label{pipeline}
\end{figure*}

\section{More Qualitative Results}

\noindent \textbf{Qualitative results of the ablation study.}
In Figure~\ref{pipeline}, we show a typical example of our method and its variants, including “w/o action triplets”, “w/ action triplets”, both “w/ action triplets” and “w/ action state descriptions”, and “w/ action-aware adaptive interaction”. 

As shown in Figure~\ref{pipeline}, compared with “w/o action triplet”, “w/ action triplet” retrieves images that depict the action of “holding”, which partially aligns with the intent of the query. Moreover, jointly using both “w/ action triplets” and “w/ action state descriptions” retrieves candidate images that align more closely with the query intent than “w/ action triplet”. This is because each action state provides more details and comprehensive descriptions,  supplementing the information contained in action triplets “\textless boy, point out, something \textgreater” and “\textless girl, hold, boy \textgreater”. However, the action knowledge of “\textless boy, point out, something \textgreater” generated by LLM contains some irrelevant action information, which misleads CLIP's fine-grained action-aware visual understanding. “w/ action-aware adaptive interaction” helps mitigate the interference from action-irrelevant contents, enabling the model to retrieve the corresponding ground truth image.

\begin{figure*}[htbp]
\centering
\begin{subfigure}[b]{14cm} 
    \centering
    \includegraphics[width=14cm, height=11cm]{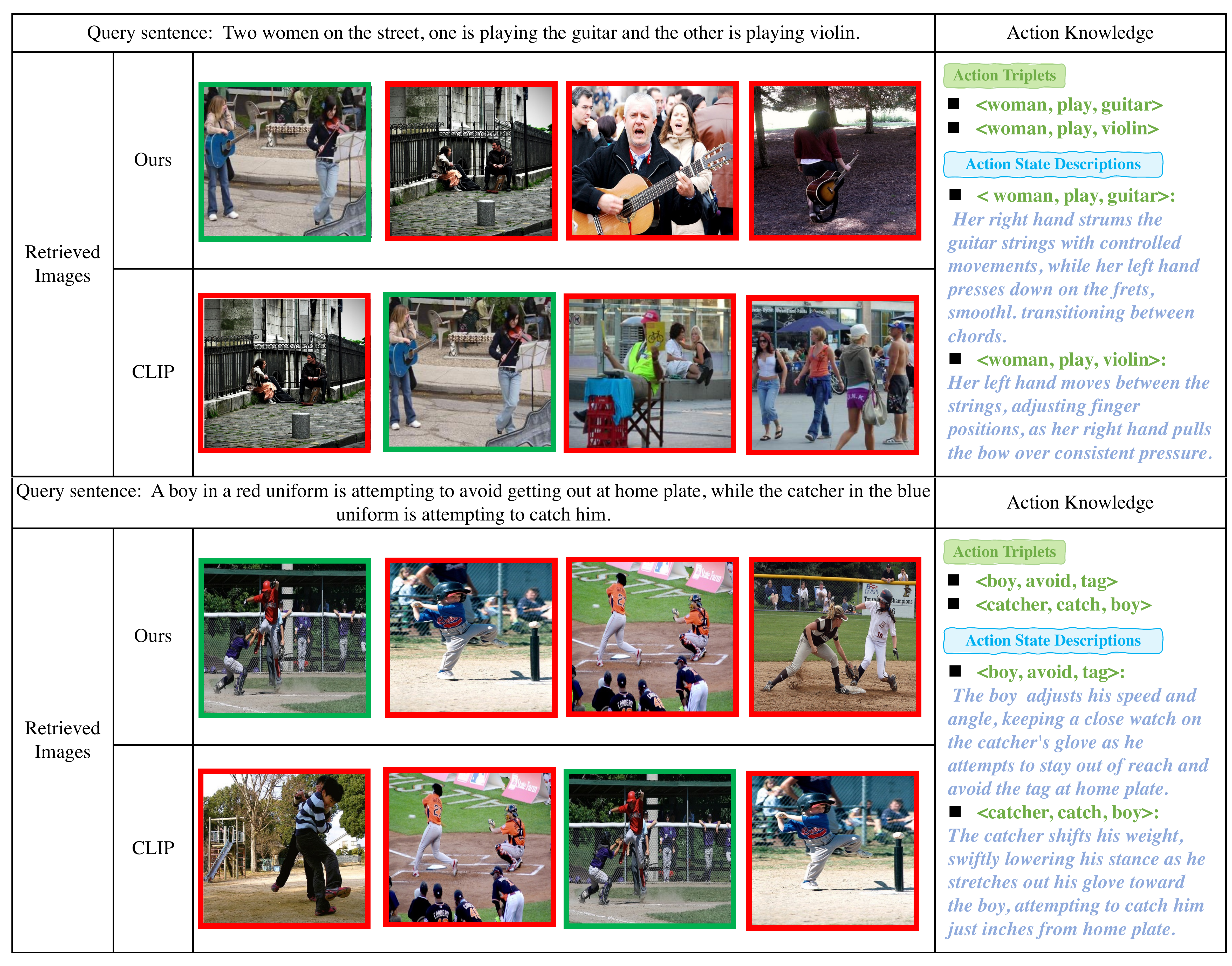}
    \caption{Examples on the Flickr30K dataset.}
    \label{fig:sub1}
\end{subfigure}
\par\bigskip 
\begin{subfigure}[b]{14cm} 
    \centering
    \includegraphics[width=14cm, height=11cm]{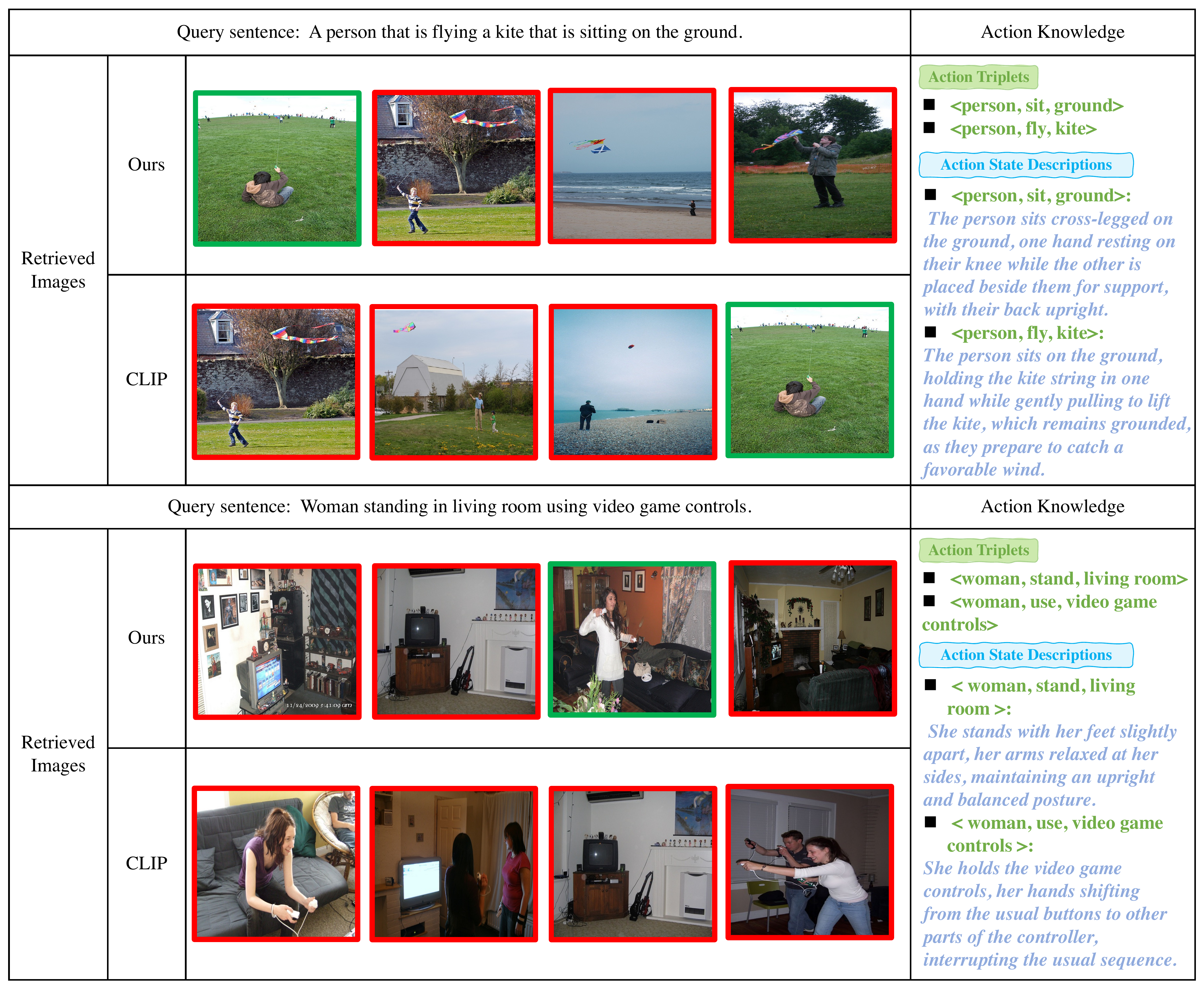}
    \caption{Examples on the COCO dataset.}
    \label{fig:sub2}
\end{subfigure}
\caption{Visual comparisons of text-to-image retrieval examples between our method and baseline CLIP on Flickr30K and COCO datasets. The ground-truth images are outlined in green boxes, and the incorrect ones are outlined in red boxes.}
\label{vis2}
\end{figure*}

\noindent \textbf{Qualitative results compared with CLIP.} To better understand the effectiveness of our method, we visualize some examples of text-to-image retrieval results on Flickr30K and COCO datasets, as shown in Figure~\ref{vis2}. For each text query, the top-4 ranked images from our method and baseline model CLIP are listed. The ground truth images are outlined in green boxes while incorrect ones are in red boxes. It can be seen that our method is more robust in complex scenes compared to CLIP, achieving better retrieval results. For example, in the two examples of Figure~\ref{vis2} (a), our method successfully retrieves the ground truth image with the precise description of “playing” and “catch”.

In the first example of Figure~\ref{vis2} (b), CLIP ranks the top-3 images incorrectly, as they do not depict the action of “sitting” from the text query. In contrast, our method can enhance the ground truth image with precise state descriptions of both “flying” and “sitting”. However, in the second example of Figure~\ref{vis2} (b), our method ranks incorrectly for a given text query, since the action knowledge generated by the LLM for “\textless woman, use, video game controls \textgreater” fails to precisely describe the action state of “use” in the original sentence context, resulting in some noise information is still retained in action-aware prompts and leading to an incorrect retrieval. Thus, using a more advanced LLM in future work may help correct this error in some content.

\end{document}